\documentclass[11pt]{article}

\usepackage[final]{acl}

\usepackage{times}
\usepackage{latexsym}

\usepackage[T1]{fontenc}
\usepackage[utf8]{inputenc}

\usepackage{setspace}
\usepackage{pifont}

\usepackage{microtype}

\usepackage{inconsolata}

\usepackage{graphicx}
\usepackage{float}

\usepackage{booktabs}
\usepackage{multirow}
\usepackage{amsmath}
\usepackage{xcolor}
\usepackage{url}
\usepackage{tikz}
\usepackage[most]{tcolorbox}
\usetikzlibrary{arrows,positioning,calc}

\definecolor{phase1color}{RGB}{54,75,154}    % #364B9A - dark blue
\definecolor{phase2color}{RGB}{152,202,225}  % #98CAE1 - light blue
\definecolor{phase3color}{RGB}{254,218,139}  % #FEDA8B - golden yellow
\definecolor{phase4color}{RGB}{246,126,75}   % #F67E4B - orange
\definecolor{phase5color}{RGB}{165,0,38}     % #A50026 - dark red
\definecolor{datacolor}{RGB}{230,230,230}    % light gray

\definecolor{userpromptframecolor}{HTML}{FF4500} % Reddit Orange for frame
\definecolor{userpromptbackcolor}{HTML}{FFF0EC}  % Very pale orange for background
\definecolor{userprompttextcolor}{HTML}{000000}  % Black for text

\definecolor{assistantpromptframecolor}{HTML}{336699} % Lapis Lazuli for frame
\definecolor{assistantpromptbackcolor}{HTML}{E8F4FC}  % Very pale blue for background
\definecolor{assistantprompttextcolor}{HTML}{000000}  % Black for text
\definecolor{darkblueheader}{HTML}{336699}            % Lapis Lazuli for header

\definecolor{redcrosscolor}{HTML}{CC0000}   % Red for X mark
\definecolor{greencheckcolor}{HTML}{008000} % Green for check mark

\definecolor{headerboxframecolor}{HTML}{FFBBAA}    % Pale orange for frame
\definecolor{headerboxbackcolor}{HTML}{FFFFFF}     % White for background
\definecolor{headerboxtitlebackcolor}{HTML}{FFBBAA} % Pale orange for title background

\newcommand{\markfont}{\fontsize{7pt}{9pt}\selectfont}

\newcommand{\incorrectmark}{{\color{redcrosscolor}\markfont\ding{55}}} % Red cross

\newcommand{\promptfont}{\ttfamily\tiny}

\tcbset{
    chatpromptbase/.style={
        fontupper=\promptfont, % Font: teletype, tiny
        boxrule=0.5pt,          % Thickness of the frame
        boxsep=1pt,             % Padding similar to \fboxsep=1pt
        left=3pt, right=3pt, top=2pt, bottom=2pt, % Additional inner padding
        before skip=0pt,        % Space before the box
        after skip=0pt,         % Space after the box
        before upper={\parindent0pt}, % No paragraph indent within the prompt
        arc=0pt
    }
}

\newtcolorbox{userprompt}[1]{
    enhanced,
    chatpromptbase,
    colback=userpromptbackcolor,
    coltext=userprompttextcolor,
    colframe=userpromptframecolor,
    coltitle=white,
    fonttitle=\bfseries\promptfont,
    colbacktitle=userpromptframecolor,
    toptitle=1pt, bottomtitle=1pt, lefttitle=2pt, righttitle=2pt,
    title={#1}
}

\newtcolorbox{assistantprompt}[1]{
    enhanced,
    chatpromptbase,
    colback=assistantpromptbackcolor,
    coltext=assistantprompttextcolor,
    colframe=assistantpromptframecolor,
    coltitle=white,
    fonttitle=\bfseries\promptfont,
    colbacktitle=darkblueheader,
    toptitle=1pt, bottomtitle=1pt, lefttitle=2pt, righttitle=2pt,
    title={#1}
}

\newtcolorbox{headerbox}[1]{
    enhanced,
    boxrule=0.3pt,
    arc=4pt,
    before skip=0pt,
    after skip=0pt,
    colframe=headerboxframecolor,
    colback=headerboxbackcolor,
    coltitle=black,
    fontupper=\promptfont,
    fonttitle=\promptfont,
    colbacktitle=headerboxtitlebackcolor,
    left=2pt, right=2pt, top=1pt, bottom=1pt,
    toptitle=0pt, bottomtitle=0pt, lefttitle=2pt, righttitle=2pt,
    title={#1}
}

\tcbset{
    phasebox/.style={
        width=0.82in,
        fontupper=\rmfamily,
        boxrule=0.8pt,
        arc=4pt,
        boxsep=2pt,
        left=2pt,
        right=2pt,
        top=2pt,
        bottom=2pt,
        colframe=black!80,
    },
    phase1/.style={phasebox, colback=phase1color, colupper=datacolor},
    phase2/.style={phasebox, colback=phase2color, colupper=black},
    phase3/.style={phasebox, colback=phase3color, colupper=black},
    phase4/.style={phasebox, colback=phase4color, colupper=black},
    phase5/.style={phasebox, colback=phase5color, colupper=datacolor},
    databox/.style={
        width=0.6in,
        fontupper=\rmfamily,
        boxrule=0.8pt,
        arc=4pt,
        boxsep=2pt,
        left=2pt,
        right=2pt,
        top=2pt,
        bottom=2pt,
        colback=datacolor,
        colframe=black!60,
        colupper=black,
    },
}

\title{\textsc{PluRule}: A Benchmark for Moderating Pluralistic Communities\\ on Social Media}
\author{
  Zoher Kachwala\textsuperscript{1}\thanks{Correspondence: \texttt{zkachwal@iu.edu}, \texttt{rasimura@iu.edu}},
  Bao Tran Truong\textsuperscript{1,2},
  Rasika Muralidharan\textsuperscript{1},\\
  \textbf{Haewoon Kwak\textsuperscript{1}},
  \textbf{Jisun An\textsuperscript{1}},
  \textbf{Filippo Menczer\textsuperscript{1}}\\
  \textsuperscript{1}Observatory on Social Media, Indiana University, USA \\
  \textsuperscript{2}Center Synergy of Systems,
  TUD Dresden University of Technology, Germany
}

\begin{document}
\maketitle

\begin{abstract}
Social media are shifting towards pluralism --- community-governed platforms where groups define their own norms. 
What violates rules in one community may be perfectly acceptable in another. 
Can AI models help moderate such pluralistic communities?
We formalize the task as a multiple-choice problem, mirroring how human moderators operate in the real world: given a comment and its surrounding context, identify which specific rule, if any, is violated. 
We introduce \textsc{PluRule}, a multimodal, multilingual benchmark for detecting 13,371 rule violations across 1,989 Reddit communities spanning 2,885 rules in 9 languages.
% For this, we introduce \textsc{PluRule}, a multimodal, multilingual benchmark consisting of 13,231 rule violations from Reddit, spanning 2,848 rules from 1,972 communities in 9 languages.
Using this benchmark, we show that state-of-the-art vision-language models struggle significantly: even GPT-5.2 with high reasoning performs only slightly better than a trivial baseline.
% Using this benchmark, we show that state-of-the-art vision-language models struggle significantly: even GPT-5.2 with high reasoning achieves only 58\% accuracy, barely 8 percentage points above a trivial baseline that always predicts no violation (50\%).
We also find that bigger models and increased context provide marginal gains, and universal rules like civility and self-promotion are easier to detect.
Our results show that moderation of pluralistic communities on social media is a fundamental challenge for language models.
Our code\footnote{\url{https://github.com/osome-iu/PluRule}} and benchmark\footnote{\url{https://hf.co/datasets/osome-iu/PluRule}} are publicly available.
\end{abstract}

\section{Introduction}
\label{sec:introduction}

\begin{figure}
  \centering
  \begin{minipage}{\columnwidth}
  \begin{userprompt}{Input Prompt}
    Subreddit Info:
    \vspace{0.5\baselineskip}
    \begin{headerbox}{Subreddit: r/santamonica: Welcome to Santa Monica, California}
    Anything related to Santa Monica, where the rent is high but the temperatures aren't.
    \end{headerbox}
    \vspace{0.5\baselineskip}
    Rules:
    \vspace{0.5\baselineskip}
    \begin{headerbox}{Rule 1:  Santa Monica-related only}
    Please keep posts to Santa Monica-specific issues. Venice and West LA issues can be posted in /r/LosAngeles
    \end{headerbox}
    \vspace{0.5\baselineskip}
    \begin{headerbox}{Rule 2:  Respect other redditors}
    Respect other r/SantaMonica users, as both individuals and as groups. If you disagree with someone here, do so gracefully \ldots
    \end{headerbox}
    \vspace{0.5\baselineskip}
    \ldots(other rules)\ldots
    \vspace{0.5\baselineskip}
    \begin{headerbox}{Rule 6:  No calling people paid shills just because they disagree with you.}
    Accusing people of being paid shills with no proof just because they disagree with you can and will result in bans.
    \end{headerbox}
    \vspace{0.5\baselineskip}
    \ldots(other rules)\ldots
    \par
    \vspace{0.5\baselineskip}
    Submission:
    \vspace{0.5\baselineskip}
    \begin{headerbox}{Homeowners association sues City and Fairmont Miramar over hotel redevelopment approval\\USER1, Tue, Dec 15, 2020, 1:37PM}
    [URL]
    \begin{center}
      \includegraphics[width=0.7\linewidth]{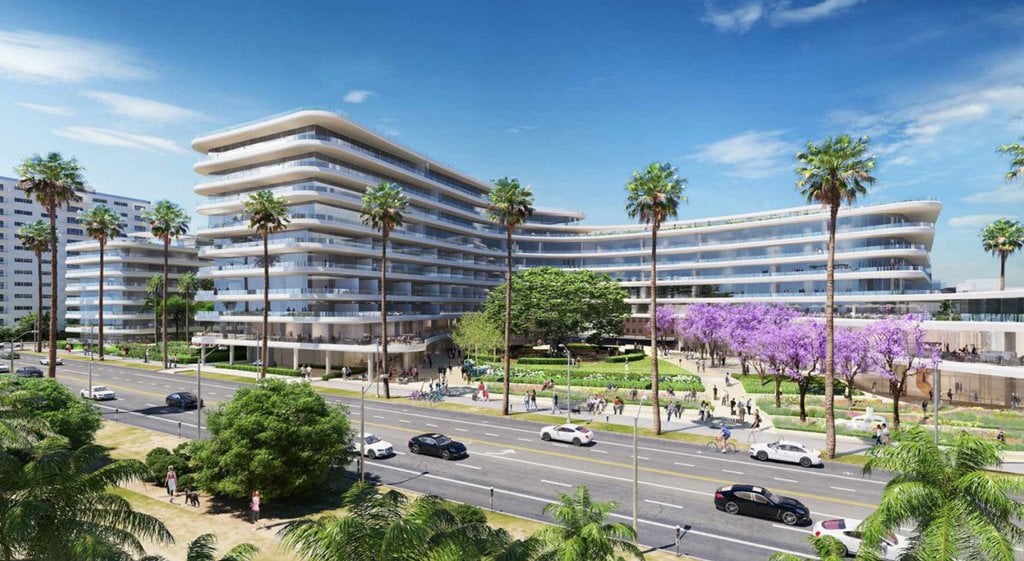}
    \end{center}
    \end{headerbox}
    \vspace{0.5\baselineskip}
    Discussion:
    \vspace{0.5\baselineskip}
    \begin{headerbox}{Comment 1: USER1, Tue, Dec 15, 2020, 1:38PM}
    And people wonder why everything is so expensive here\ldots
    \end{headerbox}
    \vspace{0.5\baselineskip}
    \ldots (other comments)\ldots
    \vspace{0.5\baselineskip}
    \begin{headerbox}{Comment 4 [TARGET COMMENT]: USER4, Mon, Dec 21, 2020, 3:58PM}
    You must work for Dell. He's been ripping off Santa Monica for years by avoiding taxes on the Fairmont Miramar. Look it up. He's a right wing a-hole
    \end{headerbox}
    \vspace{\baselineskip}
    Question: Does the [TARGET COMMENT] violate a rule?
    \vspace{0.5\baselineskip}

    (a)~Santa Monica-related only; (b)~No rules broken; (c)~Respect other redditors; (d)~Appealing bans; (e)~No calling people paid shills just because they disagree with you; (f)~No low-effort posts; (g)~No hateful speech directed at other people, including about the homeless; (h)~No spam
  \end{userprompt}
  \vspace{0.25\baselineskip}
  \begin{assistantprompt}{Model Response}
    \ldots(reasoning trace)\ldots
    \vspace{0.5\baselineskip}\\
    \textbf{Final answer:} (c) Respect other redditors. \incorrectmark
  \end{assistantprompt}
  \end{minipage}
  \caption{A \textsc{PluRule} example.
  GPT-5.2 (high reasoning) receives the full context of the target comment, then selects which rule is violated.
  Here, the correct answer is (e) but GPT-5.2 selects (c). Full text in the Appendix.}
  \label{fig:example-datapoint}
\end{figure}

% intro paragraph on why pluralistic moderation is important
Ensuring platform safety and encouraging constructive participation are among the most persistent challenges of social media governance \cite{gillespie2020content}.
On centralized platforms such as X \cite{twitter2025transparency}, YouTube \cite{youtube2024}, and Meta \cite{meta2024transparency}, content moderation is increasingly carried out by a combination of human moderators and automated detection algorithms.
These systems typically focus on narrowly defined categories --- such as unlawful content, incivility, hate speech, and harassment --- that are assumed to be universal across all users and communities.
By privileging mainstream norms, they overlook the values, languages, and forms of expression used by minority communities, leading to higher rates of content removal for marginalized groups \cite{lingel2015face, jiang2020characterizing, griffin2024heteronormative, celeste2023platform}.
Centralized platform rules therefore fail to account for the diverse experiences and contextual meanings that vary across communities \cite{DiazHechtFelella2021}.

%paragraph on the rise of decentralized social media as a result of the world's desire for pluralism
Bucking this trend, some platforms have adopted community-governed structures that allow groups to define their own norms.
Reddit, for instance, hosts hundreds of thousands of topic-based communities (subreddits), each with its own rule set in addition to platform-wide guidelines \cite{reddit_reddiquette}.
While these pluralistic structures empower communities, they also place a substantial burden on volunteer moderators.
On Reddit alone, the estimated value of this uncompensated labor exceeded \$3.4 million in 2020 \cite{li_measuring_2022}.
Unsurprisingly, moderators are often eager to adopt automated tools that can reduce their burden \cite{CJR, dosono_moderation_2019, hill_how_2019, lloyd_there_2025}.

However, the contextual nature of community-specific rules poses a fundamental challenge for automation.
What violates a rule in one community may be perfectly acceptable in another \cite{chandrasekharan2019crossmod, li_all_2022}.
A satirical insult about someone's appearance, for instance, is encouraged in r/RoastMe but would violate civility rules in most other communities.
Similarly, self-promotion that constitutes spam in most subreddits is required in creative showcase communities.
Effective moderation requires understanding not just the rule text, but the implicit norms, values, and purposes that each community has developed over time.

%paragraph on why pluralistic moderation with AI important
Given these contextual complexities, the question arises whether modern AI systems can effectively assist with pluralistic moderation.
The central challenge is whether language models can recognize that identical content may be acceptable in one community but violate rules in another.
Even similar rules may be interpreted differently depending on local community norms \cite{selbst2019fairness, birhane2021multimodal}.  
% Most AI moderation systems focus on universal violations like toxicity, hate speech, or spam, while community-based moderation operates at the local level where each group defines its own norms.

% paragraph on task desription and the first contribution
To investigate this question empirically, we formalize the detection of rule violations as a multiple-choice task that mirrors how human moderators operate in practice (Figure~\ref{fig:example-datapoint}). 
We introduce \textsc{PluRule}, the first multimodal, multilingual benchmark for moderating pluralistic communities on social media.
The benchmark comprises 13,371 moderation instances with 72,675 comments and 3,643 images, spanning 1,989 subreddits with 2,885 distinct rules across 9 languages.
\textsc{PluRule} incorporates substantial diversity along two dimensions: 25 semantically-derived subreddit categories (e.g., politics, gaming, music) and 27 rule categories (e.g., civility, self-promotion, spoilers).
% Each instance is structured as a contrastive pair: one comment that violates a community rule and one compliant comment from the same submission.
%Our benchmark is pluralistic across two dimensions: 27 categories of subreddits, and 31 categories of rules.

% paragraph on second contribubtion: i.e. results
Using \textsc{PluRule}, we evaluate state-of-the-art vision-language models (VLMs) on the detection of rule violations under different context conditions.
%: the comment alone, the full discussion thread, the original submission, user identities, and associated images.
Our results reveal substantial limitations: even GPT-5.2 with high reasoning effort achieves only 58\% accuracy, barely exceeding a trivial baseline that always predicts no violation (50\%). 
Providing additional context --- the discussion thread, original submission, participant labels, and images --- improves GPT-5.2's performance by only 2--3 percentage points.
Open-weight models like Qwen3-VL-Instruct and Qwen3-VL-Thinking perform even worse, failing to surpass baseline performance.
Performance breakdown by rule category reveals that models successfully detect universal violations such as civility (69\%) and self-promotion (63\%), but fail on rules that require contextual understanding; low-effort (43\%), evidence-based (47\%), and relevance (44\%) all fall below baseline. 
These results reveal a critical gap: current VLMs can enforce universal norms but cannot adapt to the diverse, context-dependent standards that define pluralistic moderation.

\section{Related Work}
\label{sec:related-work}

% Need to move from the dectection of universal norms to pluralistic values 
Existing datasets for content moderation focus on narrow categories such as toxic speech \cite{hoang_toxcl_2024}, hate speech \cite{nghiem_hatecot_2024}, or misogyny \cite{sheppard_biasly_2024}.
Automated systems trained on these datasets are limited to detecting broadly unacceptable content under singular global standards of appropriateness. 
% , and fail to model how moderation decisions depend on community-specific rules and local context.
This assumption breaks down in decentralized platforms, where different demographic groups significantly diverge about what is considered respectful, emotionally appropriate, or toxic \cite{sachdeva2022measuring,ali2025operationalizing}. 
Moderation on such platforms must account for pluralism. 
On Reddit, for example, rules extend beyond toxicity \cite{binns_like_2017, matias_civic_2019} to include locally defined norms around formatting, tone, and ideological or topical relevance \cite{chandrasekharan2019hybrid}. 

% Subjective and interpretive nature of Reddit moderation work 
Even the enforcement of similar norms can vary widely across communities \cite{chandrasekharan2018internet}. 
% This variability highlights the importance of capturing not only which rules apply, but also the context in which a potential violation occurs, including linguistic cues \cite{blodgett_demographic_2016} and multimodal content \cite{gomez_exploring_2020, kiela_hateful_2020}. 
On Reddit in particular, moderators routinely interpret rules and assess the appropriateness of content relative to local community values rather than mechanically executing fixed policies \cite{li_all_2022,fiesler_reddit_2018, matias_civic_2019}. 
Consequently, within a community, multiple moderators can diverge when guidelines are broad or context-dependent \cite{binns_like_2017, chandrasekharan2019crossmod}. 
Across communities, the same content may be acceptable in one context while violating norms in another --- a distinction that models trained on aggregated data from multiple communities often fail to capture \cite{sap-etal-2022-annotators, Rajietal_aigap}. 
% Recent work has expanded this line of work through LLMs, however, several works emphasize toxicity and targeted speech \cite{Kumar_AbuHashem_Durumeric_2024, Mirko_LLM}. This leaves our understanding of pluralistic and community-specific moderation nature of moderation --- such as that of Reddit ---remains underexplored.

%Additionally, studies expanding moderation studies using LLMs emphasize toxicity and targeted speech \cite{Kumar_AbuHashem_Durumeric_2024, Mirko_LLM}. This view limits our understanding of how automated tools can be applied to real-world moderation systems. 

Previous work attempts to model the community-dependent nuance of moderation, but does not address its context-dependent nature. \citet{chandrasekharan2019hybrid} identify a small set of recurring ``macro'' norms shared across communities. 
\citet{park_detecting_2021} introduce a text-only dataset that collapses thousands of community-specific rules into coarse-grained types. 
This approach abstracts thousands of individual subreddit rules into a limited number of universal categories, obscuring differences that define each community. 
\cite{he_cpl-novid_2024} provide models with individual rules for binary yes/no judgments. 
% This formulation also prevents evaluation of whether models can distinguish between similar but distinct rules within the same community---a critical capability for real-world moderation.

% Positioning PluRule
% ??? In practice, this involves examining a comment within its full conversational context, including preceding comments or attached images, and selecting the relevant rule from the set of community-defined options. 
\textsc{PluRule} advances beyond prior work along three key dimensions. 
First, it explicitly models pluralism: instead of applying a fixed set of universal categories, models must reason over distinct, community-defined rules.
Second, it frames moderation as a rule identification task (multiple-choice) rather than binary classification. This mirrors real-world moderator workflows and enables more fine-grained evaluation.
Finally, \textsc{PluRule} is multilingual and multimodal, capturing the visual \cite{gomez_exploring_2020} and linguistic \cite{blodgett_demographic_2016} diversity of online communities often overlooked by text-only benchmarks.

\section{\textsc{PluRule} Benchmark}

% \textsc{PluRule} is designed to evaluate the core challenge of pluralistic alignment: can a model correctly identify that the same content violates different rules (or no rules at all) depending on community context?
% The benchmark is multimodal and multilingual, reflecting the reality that moderation decisions depend on visual context \cite{gomez_exploring_2020, kiela_hateful_2020} and occur in diverse linguistic communities \cite{blodgett_demographic_2016}. 

%\paragraph{Task.}
\textsc{PluRule} formalizes the task of moderating pluralistic communities on Reddit as a multiple-choice question (Figure~\ref{fig:example-datapoint}).
Given a comment from a specific community (subreddit), models must identify which specific rule, if any, has been violated.

%\paragraph{Input Format.}
For each comment, models receive the community's rules along with the surrounding context that moderators consider when making decisions.
The context includes: (1)~the discussion thread that precedes the comment; (2)~the submission post to which the comment responds, including any images; and (3)~anonymized identifiers of the participants in the discussion.

%\paragraph{Instance Structure.}
Each \textit{moderation instance} in \textsc{PluRule} consists of a pair: a violating comment and a compliant comment with overlapping context from the same submission.
Models are evaluated on both comments separately.
For both comments, models are presented with answer options consisting of all subreddit rules plus a ``No rules broken'' option, labeled (a), (b), (c), etc.
Each comment's answer options are deterministically shuffled using a seed based on the comment ID to prevent models from exploiting positional bias. 
The correct answer for violating comments is the violated rule; for compliant comments, it is ``No rules broken.''
Since half the comments violate a rule and half do not, always predicting ``No rules broken'' yields a majority baseline of 50\% accuracy. 

\section{\textsc{PluRule} Construction}
\label{sec:pipeline}

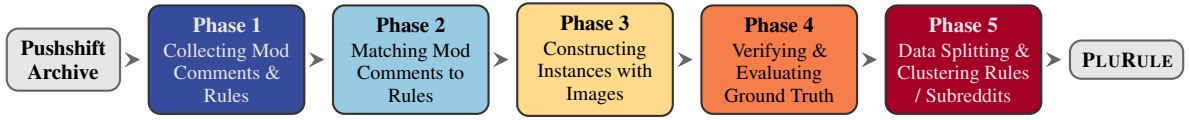
\begin{figure*}
\centering
\newcommand{\headerfontsize}{\fontsize{8}{9}\selectfont}
\newcommand{\bodyfontsize}{\fontsize{7.5}{8.5}\selectfont}
\begin{tikzpicture}[
    node distance=0.08cm,
    arrow/.style={-stealth, line width=1.5pt, draw=black!60, shorten >=-1pt}
]

% Input
\node[name=input] {
    \begin{tcolorbox}[databox]
        \centering\headerfontsize\textbf{Pushshift Archive}
    \end{tcolorbox}
};

% Phase 1
\node[name=p1, right=of input] {
    \begin{tcolorbox}[phase1]
        \centering\headerfontsize\textbf{Phase 1} \\[1pt]
        \centering\bodyfontsize
        Collecting Mod Comments \& Rules
    \end{tcolorbox}
};

% Phase 2
\node[name=p2, right=of p1] {
    \begin{tcolorbox}[phase2]
        \centering\headerfontsize\textbf{Phase 2} \\[1pt]
        \centering\bodyfontsize
        Matching Mod Comments to Rules
    \end{tcolorbox}
};

% Phase 3
\node[name=p3, right=of p2] {
    \begin{tcolorbox}[phase3]
        \centering\headerfontsize\textbf{Phase 3} \\[1pt]
        \centering\bodyfontsize
        Constructing Instances with Images
    \end{tcolorbox}
};

% Phase 4
\node[name=p4, right=of p3] {
    \begin{tcolorbox}[phase4]
        \centering\headerfontsize\textbf{Phase 4} \\[1pt]
        \centering\bodyfontsize
        Verifying \& Evaluating Ground Truth
    \end{tcolorbox}
};

% Phase 5
\node[name=p5, right=of p4] {
    \begin{tcolorbox}[phase5]
        \centering\headerfontsize\textbf{Phase 5} \\[1pt]
        \centering\bodyfontsize
        Data Splitting \& Clustering Rules / Subreddits
    \end{tcolorbox}
};

% Output
\node[name=output, right=of p5] {
    \begin{tcolorbox}[databox, width=0.6in]
        \centering\headerfontsize\textbf{\textsc{PluRule}}
    \end{tcolorbox}
};

% Arrows
\draw[arrow] (input.east) -- (p1.west);
\draw[arrow] (p1.east) -- (p2.west);
\draw[arrow] (p2.east) -- (p3.west);
\draw[arrow] (p3.east) -- (p4.west);
\draw[arrow] (p4.east) -- (p5.west);
\draw[arrow] (p5.east) -- (output.west);

\end{tikzpicture}
\caption{\textsc{PluRule} construction pipeline}
\label{fig:pipeline}
\end{figure*}

We select Reddit as a platform because moderation actions are public: a moderation action occurs when a human moderator leaves a comment explaining a rule violation (e.g., ``Your comment violates Rule 2'').
We construct \textsc{PluRule} by starting from such moderator comments in the Pushshift Reddit archives \cite{baumgartner_pushshift_2020} and transforming them into structured benchmark instances with verified rule labels, contrastive pairs, and semantic clustering. 
Below we describe the five-phase pipeline for this construction process.

\subsection{Phase 1: Data Collection}

%\paragraph{Moderator Comments.}
We start from a publicly hosted, extended version of the Pushshift Reddit archives \cite{cohen_academic_2014}, containing approximately 15 billion comments across 40 thousand subreddits.
From these archives, we extract comments by moderators, flagged by a ``distinguished'' field in the comment object.  
To focus only on comment (not post) violations, we exclude top-level replies to submission posts. 
We filter out accounts with usernames that match bot-related keywords, e.g., ``bot,'' ``automod.''
This yields approximately 10 million moderator comments across 40 thousand subreddits.

%\paragraph{Community Rules.}
We then query the Reddit API to collect each subreddit's full object.
This helps to retrieve its current rules, infer its official language, and determine if it is NSFW (not-safe-for-work).
Each subreddit must have at least one moderator comment and at least two explicit rules.
We also exclude communities with adult content (NSFW). 
After filtering for these criteria, we obtain 17,468 subreddits with 131,400 rules and approximately 9 million moderator comments.

\subsection{Phase 2: Rule Matching}

Moderators often reference rules in their comments (e.g., ``Rule 3 -- No personal attacks''). 
Despite this, linking historical moderator comments to specific rule violations is a key challenge because rules evolve over time: new ones are added, old ones are deleted, and numbering and wording of existing rules can change. 
%Thus, exact matching to rules can be inaccurate. 
%\paragraph{Semantic Similarity.}
Since the Reddit API provides only present-day rules (as of November 2025), we match the full text of moderator comments to the full text of current rules.

We use Qwen3-Embedding-8B, a multilingual text embedding model, to encode all 9 million moderator comments and 131,400 rules as dense vectors.
For each comment, we compute cosine similarity against all rules of the subreddit to which it belongs (7.5 rules on average), producing 90 million comment-rule scores.

%\paragraph{Threshold Filtering.}
We apply two thresholds to infer high-quality labels.
First, a \textit{match threshold} at the 99.2nd percentile of similarity scores (0.79): only comment-rule pairs above this threshold count as matches.
Second, an \textit{ambiguity threshold} at the 98th percentile (0.75): if multiple rules for a single comment exceed this threshold, we discard the comment entirely to avoid inferring ambiguous labels.
The 9 million moderator comments yielded 174,412 ambiguous cases and 672,493 matched comments.
% We defer verification of matches to Phase~4 after additional filtering.
In Phase~4, after additional filtering, we verify the quality of these matches.

\begin{figure*}
  \centering
  \includegraphics[width=\textwidth]{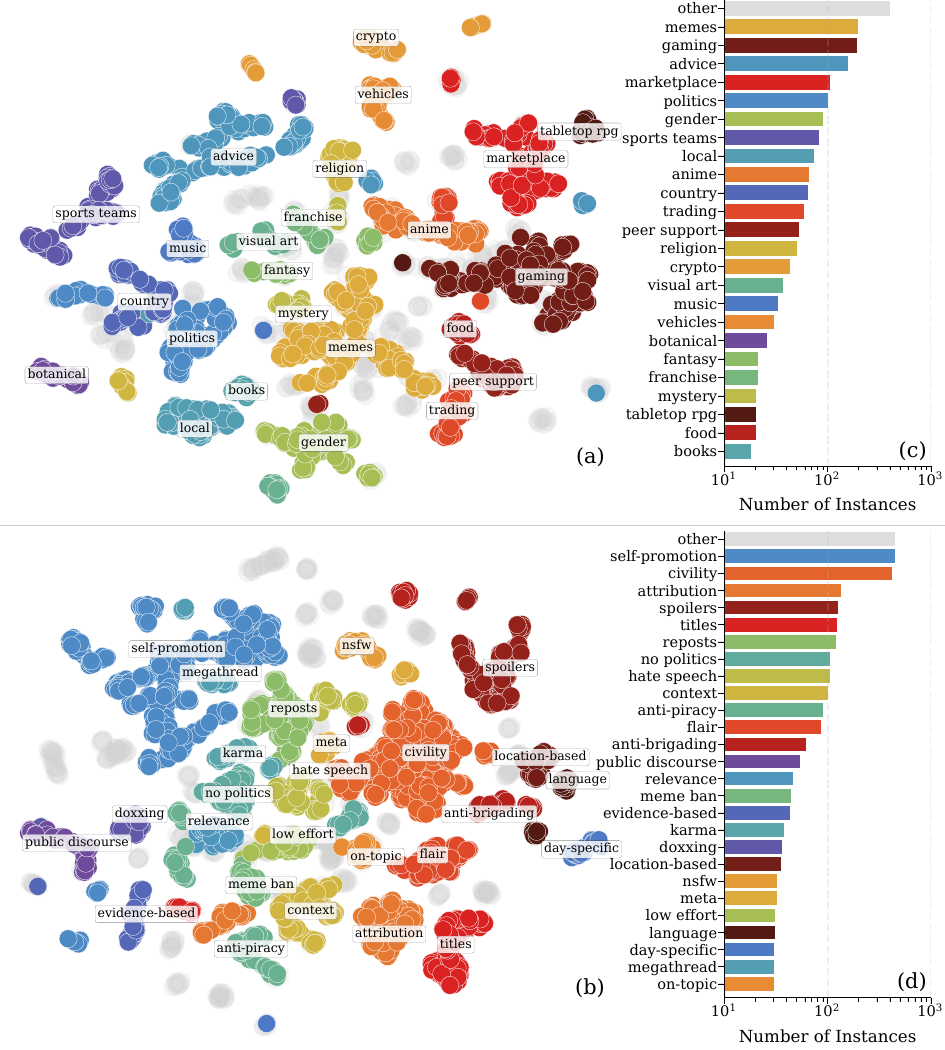}
  \caption{Diverse communities and rules in the \textsc{PluRule} benchmark. Left: 2D UMAP visualizations of (a)~1,989 subreddits and (b)~2,885 rules, with colors indicating cluster assignments by HDBSCAN. Grey points represent unclustered items categorized as ``other''. Right: Distributions of 13,371 instances across (c)~25 subreddit clusters and (d)~27 rule clusters, with bar colors matching the clusters.}
  \label{fig:clusters}
\end{figure*}

\begin{table*}
  \centering
  \setlength{\tabcolsep}{4pt}
  \begin{tabular}{lrrrrrr}
  \toprule
  \textbf{Split} & \textbf{Instances} & \textbf{Comments} & \textbf{Images} & \textbf{Subreddits / Clusters} & \textbf{Rules / Clusters} & \textbf{Languages} \\
  \midrule
  Train & 9,155 & 51,968 & 2,077 & 861 / 25 & 1,336 / 27 & 9 \\
  Val & 1,382 & 7,631 & 376 & 537 / 25 & 586 / 27 & 9 \\
  Test & 2,834 & 13,076 & 1,190 & 1,989 / 25 & 2,039 / 27 & 9 \\
  \midrule
  \textbf{Total} & \textbf{13,371} & \textbf{72,675} & \textbf{3,643} & \textbf{1,989 / 25} & \textbf{2,885 / 27} & \textbf{9} \\
  \bottomrule
  \end{tabular}
  \caption{\textsc{PluRule} statistics. Each instance contains one rule-violating and one compliant thread from the same submission. All 1,989 subreddits appear in the test set. We count only the 9 languages with at least 10 instances each. There are 22 languages with at least one instance.}
  \label{tab:dataset-stats}
\end{table*}

\subsection{Phase 3: Instance Construction}

%\paragraph{Building Comment Trees.}
We wish to capture the complete conversational context of each violation, i.e., the full comment thread leading to the rule-violating comment --- the one to which the moderator replied. 
We collect all comments from the same submission by matching submission IDs in the Pushshift archives, yielding 73.8 million comments.
We then build comment trees representing the reply structure.
From each tree, we extract a \textit{violating thread}: the path up from the rule-violating comment, through its parent comments, to the root submission.

Effective moderation requires a capability to discriminate between similar rule-violating and compliant comments within the same discussion. 
To this end, we create a \textit{moderation instance} by pairing each violating thread with a \textit{compliant thread} --- a discussion branch from the same submission that received no moderator action. 
We first collect candidate compliant threads whose leaf comment lies at either depth $n$ (same as the violating thread) or depth $n-1$. 

We then apply six filtering criteria.
For both violating and compliant threads, we exclude: (1)~deleted/removed content or deleted users to ensure complete discussion context; (2)~media in comments to limit images to submissions only; and (3)~any moderator-authored comments in the thread to avoid back and forth discussions between a moderator and a user.
For violating threads specifically, we exclude (4)~edited leaf comments that became compliant after moderator intervention.
For compliant threads specifically, we exclude (5)~leaf comments with moderator replies to ensure no moderator flagged these comments as violations.
At the instance level, we further exclude (6)~instances whose moderator comment was posted on or after March~1, 2023 to adhere to the Pushshift coverage window.

After filtering, we rank candidate compliant threads for each violating thread using three criteria to maximize the shared context between the two: 
(1)~higher number of common ancestors; (2)~higher thread depth to prioritize $n$ over $n-1$; and (3)~lower vote score, to select less popular content that nevertheless complied with community rules. 
We select the compliant thread that ranks highest. 

To complete moderation instances with the context from which the discussions originated, for each thread pair we collect the corresponding submission objects from Pushshift.
We filter out instances whose submissions contain NSFW content, crossposts, or videos.
For the remaining submissions, we download images using a priority hierarchy: gallery images, direct URLs, video thumbnails, and Reddit-cached previews as fallbacks. 
Each download is validated for image content type and capped at 50 MB. 
We also exclude submissions with deleted/removed content, deleted users and those posted by moderators.

From 672,493 matched comments, we build 378,334 comment trees and successfully create 16,289 instances, totaling 32,578 threads.
Most failures for threads and submissions stem from deleted/removed content or deleted users.

\subsection{Phase 4: Verification}

%\paragraph{LLM Verification.}
We use a large language model, Qwen3-30B-A3B-Instruct, to verify the rule matches inferred in Phase~2 for the benchmark instances. 
For each instance, we present the model with the \textit{moderator comment} and \textit{matched rule} to classify the comment as: (a)~stating a violation of the rule, (b)~discussing the rule, or (c)~unrelated to the rule. 
We retain as ground-truth labels only matched rules classified as (a), achieving an 82.1\% verification rate (13,371 out of 16,289 instances).
This step filters out incorrect matches and cases where moderators mentioned a rule without enforcing it.

To evaluate the accuracy of these ground-truth labels, three authors independently annotated 100 moderator comments sampled randomly from English subreddits.
For each moderator comment, annotators selected which subreddit rule was violated from the available options --- the same task performed by the matching pipeline.
For 85 of the comments in the sample, all three annotators agreed on the label. In 12 cases, the label was assigned based on a majority (two of the three annotators agreed). In the 3 remaining cases, there was no majority agreement and the label was adjudicated after further inspection by one annotator.
Comparing the pipeline's labels against this human-established ground truth, we found 96\% overall accuracy: 100\% on full-agreement cases (85/85), 66.67\% on majority-agreement cases (8/12), and 100\% on adjudicated cases (3/3).

\subsection{Phase 5: Data Splitting and Clustering}

%\paragraph{Train/Validation/Test Splits.}
We split the instances into training, validation, and test sets using a strategy based on the number of instances per subreddit. 
For subreddits with a single instance, we allocate the instance to the test set. 
For subreddits with two instances, we allocate one instance to the training set and one to the test set. 
For subreddits with 3--9 instances, we allocate one each to test and validation sets, and the remaining to the training set.  
For subreddits with 10 or more instances, we use a 80/10/10 split for the training, validation, and test sets.
This ensures all communities appear in the test set while preventing any single community from dominating the evaluation.

%\paragraph{Clustering Subreddits and Rules.}
To analyze model accuracy across communities, we cluster subreddits and rules based on their semantic embeddings. 
For subreddits, we embed the subreddit name, title, and description. 
For rules, we embed the concatenation of short name, description, and violation reason.  
We apply UMAP for dimensionality reduction using cosine distance on the 4,096-dimensional Qwen3-Embedding-8B, then HDBSCAN for density-based clustering (see Appendix~\ref{sec:app:methods}). 
We visualize the resulting clusters and their distribution in Figure~\ref{fig:clusters}.

We labeled each cluster using Qwen3-30B-A3B-Thinking (see Appendix~\ref{sec:app:methods}) followed by manual refinement. 
We assigned these cluster labels to all instances, enabling both fine-grained and category-level evaluation.
Table~\ref{tab:dataset-stats} provides full statistics of the \textsc{PluRule} dataset. 
%3,692 distinct rules across 2,419 subreddits. 

% \subsection{Community Diversity}
% - Types of communities represented
% - Community size distribution
% - Topical diversity analysis

% \subsection{Rule Categories and Distribution}
% - Rule taxonomy development
% - Rule complexity analysis
% - Violation distribution per subreddit
% - Cross-community rule patterns

% \subsection{Content Modality Analysis}
% - Text-only vs multimodal instances
% - Image content characteristics
% - Discussion thread depth statistics

% \subsection{Temporal and Contextual Properties}
% - Thread reconstruction completeness
% - Context window analysis
% - Data quality metrics

\section{Evaluation}
\label{sec:evaluation}

\subsection{Experimental Setup}
\label{sec:experimental-setup}

For each instance, models receive the subreddit description, complete rule set, and surrounding context, then select the correct answer from the multiple-choice options.
We report accuracy on the test set, with the 50\% baseline corresponding to always predicting ``No rules broken.''
We compute 95\% confidence intervals via bootstrap resampling with 100 thousand iterations.
%, using shared resample indices across all models to ensure comparable intervals.

\begin{table*}[t]
  \centering
  \setlength{\tabcolsep}{3.75pt}
  \begin{tabular}{lllllllll}
  \toprule
  \textbf{Models} & \multicolumn{2}{c}{\textbf{Qwen3-VL-4B}} & \multicolumn{2}{c}{\textbf{Qwen3-VL-8B}} & \multicolumn{2}{c}{\textbf{Qwen3-VL-30B}} & \multicolumn{2}{c}{\textbf{GPT-5.2}} \\
  \cmidrule(lr){2-3} \cmidrule(lr){4-5} \cmidrule(lr){6-7} \cmidrule(lr){8-9}
  \textbf{Variants} & \multicolumn{1}{c}{Instruct} & \multicolumn{1}{c}{Thinking} & \multicolumn{1}{c}{Instruct} & \multicolumn{1}{c}{Thinking} & \multicolumn{1}{c}{Instruct} & \multicolumn{1}{c}{Thinking} & \multicolumn{1}{c}{Low} & \multicolumn{1}{c}{High} \\
  \midrule
  Comment Only & \textbf{49.6} & 37.4 & \textbf{51.0} & 40.3 & 50.2 & 46.1 & 54.1 & 55.0 \\
  \quad +Discussion & 49.2 {\tiny (-0.4)} & 39.8 {\tiny (+2.4)} & 50.7 {\tiny (-0.3)} & 43.9 {\tiny (+3.6)} & 51.0 {\tiny (+0.8)} & 48.2 {\tiny (+2.1)} & 55.3 {\tiny (+1.2)} & 56.2 {\tiny (+1.2)} \\
  \quad\quad +Submission & 48.3 {\tiny (-0.9)} & 44.9 {\tiny (+5.1)} & 49.2 {\tiny (-1.5)} & \textbf{47.2} {\tiny (+3.3)} & 51.1 {\tiny (+0.1)} & 49.1 {\tiny (+0.9)} & 56.8 {\tiny (+1.5)} & 57.3 {\tiny (+1.1)} \\
  \quad\quad\quad +User & 48.9 {\tiny (+0.6)} & \textbf{45.0} {\tiny (+0.1)} & 50.0 {\tiny (+0.8)} & 46.7 {\tiny (-0.5)} & \textbf{52.4} {\tiny (+1.3)} & 49.4 {\tiny (+0.3)} & \textbf{57.4} {\tiny (+0.6)} & \textbf{57.7} {\tiny (+0.4)} \\
  \quad\quad\quad\quad +Images & 48.4 {\tiny (-0.5)} & 45.0 {\tiny (+0.0)} & 49.8 {\tiny (-0.2)} & 44.9 {\tiny (-1.8)} & 52.3 {\tiny (-0.1)} & \textbf{49.5} {\tiny (+0.1)} & \textbf{57.4} {\tiny (+0.0)} & 57.6 {\tiny (-0.1)} \\
  \midrule
  Baseline & \multicolumn{8}{c}{50.0} \\
  \bottomrule
  \end{tabular}
  \caption{Accuracy (\%) across models and context levels. Numbers in parentheses show differences from the previous row. Best-performing contexts for each model variant are highlighted in bold. 95\% CI for all values do not exceed $\pm 1.3\%$. The baseline corresponds to always predicting ``no rules broken.''} 
  \label{tab:results-across-contexts}
\end{table*}

\begin{figure*}
  \centering
  \includegraphics[width=\textwidth]{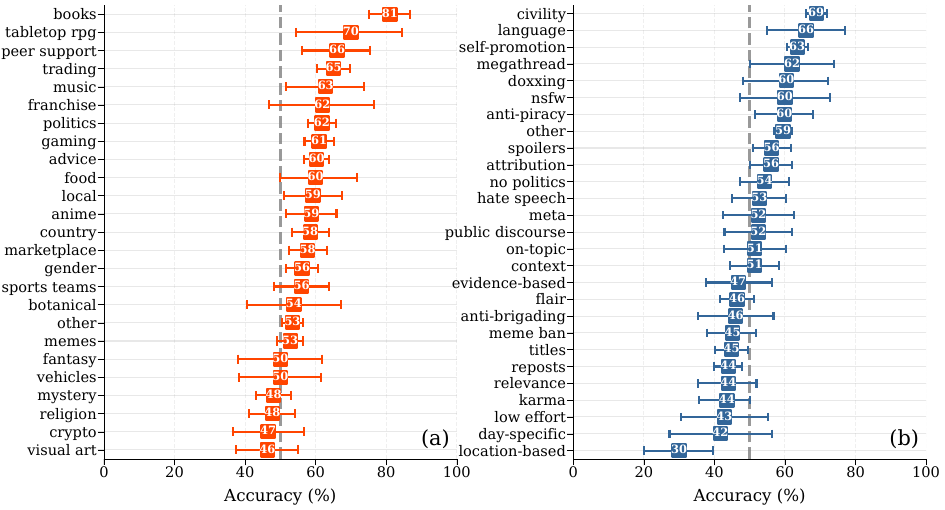}
  \caption{Accuracy for GPT-5.2 (high reasoning) with full context by (a)~subreddit clusters and (b)~rule clusters. Error bars show 95\% CI. Dashed lines indicate the baseline. Results for other models are in Appendix~\ref{sec:app:results}.}
  \label{fig:baseline-accuracy}
\end{figure*}

We evaluate three open-weight Vision-Language Models from the Qwen3-VL family for their diversity in sizes (4B, 8B, and 30B) and OpenAI's flagship model GPT-5.2. 
For each Qwen model, we test both Instruct and Thinking variants.
For GPT-5.2, we test with low and high reasoning effort.
Qwen models use temperature 0 and seed 0 for reproducibility.

We use a two-stage evaluation pipeline. In Stage~1, the model generates a free-form response to the input (see Fig.~\ref{fig:example-datapoint}). In Stage~2, we append ``Final Choice:'' to prompt the model for its final answer, then extract the selected option (a--h) using a regular expression.
For GPT-5.2, Stage~2 uses Qwen3-VL-30B-Instruct for answer extraction.

To understand which contextual signals aid rule violation detection, we test five cumulative context levels, where each level adds information to the previous: (1) \textbf{Comment Only} --- the target comment; (2) \textbf{+ Discussion} --- the full comment thread leading to the target comment; (3) \textbf{+ Submission} --- title and body text of the original post that initiated the discussion; (4) \textbf{+ User} --- anonymized author labels (USER1, USER2, etc.) to track participants; and (5) \textbf{+ Images} --- media from the submission, when available.
All levels include the subreddit description and complete rule set as baseline.
We further analyze performance by subreddit cluster and rule cluster to identify which community types and rule categories pose the greatest challenges.

\subsection{Results}
\label{sec:results}

Table~\ref{tab:results-across-contexts} reports accuracy across model sizes, context configurations, and reasoning variants (see Appendix~\ref{sec:app:results} for violating/compliant comment breakdowns.)
GPT-5.2 substantially outperforms all Qwen variants, with high reasoning effort achieving 57.7\% accuracy --- nearly 8 points above the 50\% baseline --- while Qwen models barely exceed the baseline regardless of scale.
If we were to weigh violating comments more heavily than compliant ones, the results are worse across all models (see Appendix~\ref{sec:app:results}).

Focusing on GPT-5.2 with high reasoning, additional context provides limited signal for rule violation detection: using the full context improves performance by only 2.7 percentage points above comment-only, with discussion comments yielding the largest incremental gain (+1.2 points).
%The complete discussion context provides limited signal for rule violation detection. 

Extended reasoning does not necessarily help with this task: Qwen Thinking variants underperform their Instruct counterparts, while the differences between GPT-5.2 with low and high reasoning are not significant. 

Figure~\ref{fig:baseline-accuracy} breaks down accuracy by subreddit and rule clusters for GPT-5.2 (high reasoning) using full context. 
See Appendix~\ref{sec:app:results} for breakdowns by violating/compliant comments and languages. 
Performance varies substantially across community and rule clusters.
For example, civility (69\%), language (66\%), and self-promotion (63\%) rules are detected reliably, but low effort (43\%), relevance (44\%), and evidence-based (47\%) perform worse than baseline.
To help interpret the higher accuracy in some rule clusters, we examined the number of rules in these clusters and the number of subreddits that contain these rules.
As reported in Appendix~\ref{sec:app:results}, civility and self-promotion contain far more rules than any other cluster, and those rules appear in far more subreddits than rules from other clusters.
This suggests that civility and self-promotion are universal rule types, and that models are better at detecting violations of such rules than rules requiring local context.
We report similar results obtained with other models in Appendix~\ref{sec:app:results}, with additional breakdowns by violating/compliant comments and languages.

\section{Discussion}
\label{sec:discussion}

\textsc{PluRule} provides a testbed for measuring progress toward models that can help moderate diverse community standards.
Unlike prior datasets that use coarse-grained categories or single-rule binary classification, \textsc{PluRule} requires models to distinguish among all of a community's rules simultaneously --- mirroring the decision space faced by human moderators.
We evaluate whether a single model can serve as an expert moderator across thousands of communities.

Our results reveal a gap: models succeed on universal violations like civility and self-promotion, but struggle with rules that vary across communities.
The two-dimensional variability --- 46--81\% accuracy across community types and 30--69\% across rule categories --- suggests that VLMs may be internalizing universal standards from training.

Beyond content moderation, \textsc{PluRule} evaluates whether AI systems can respect diverse human communities rather than imposing uniform standards.
Fine-tuning on community-specific examples might help models learn local norms, though scalability is a challenge.
Retrieval-augmented approaches that condition on historical moderation decisions offer another promising direction.
The semantic clustering we provide enables analysis of transfer learning: can models trained on one community or rule type generalize to similar ones?

We release \textsc{PluRule} in dehydrated form --- only IDs and our derived labels --- along with scripts to rehydrate the content from the Pushshift archives and rebuild the benchmark end-to-end.

\section{Limitations}
\label{sec:limitations}

\textsc{PluRule} was constructed from publicly available data where moderators left comments citing rule violations.
Private moderator communications, removed content, and shadow-banned posts are not accessible, meaning communities that moderate silently are underrepresented.
This likely biases the dataset toward less severe violations, as serious offenses are often removed without comment.
English-language subreddits dominate due to Reddit's user demographics; findings may not generalize to platforms with different community structures or moderation practices.

Our pipeline matches historical moderator comments (2005--2023) to rule sets retrieved in November 2025.
While semantic matching handles rule rewording and renumbering, it cannot account for rules that were added, removed, or fundamentally changed over time.
Some matches may therefore be anachronistic.

Finally, certain violations require information unavailable in our dataset.
Detecting ban evasion or repeat offenders requires historical user data that we do not collect for privacy reasons.

\section{Ethical Considerations}
\label{sec:ethics}

% \paragraph{Privacy.}
In the \textsc{PluRule} dataset, all usernames are anonymized by generic labels (USER1, USER2, etc.).
% We do not collect or release any private user metadata such as IP addresses, email addresses, or account history.

% \paragraph{Potential for misuse.}
Models trained on \textsc{PluRule} could potentially be misused to evade moderation by learning what content triggers enforcement.
This risk is inherent to any dataset that captures moderation decisions.
We believe the research benefits outweigh this risk, as understanding moderation patterns is essential for developing robust systems.

%% Uncomment when accepted 
\section*{Acknowledgments}
We gratefully acknowledge NVIDIA for the GPU access that made this study possible. 
Z.K. received support from the Institute for Humane Studies under grant nos.IHS018765.
H.K. was supported by the Republic of Korea's MSIT (Ministry of Science and ICT), under the Global Research Support Program in the Digital Field Program (RS-2024-00425354) supervised by the IITP (Institute of Information and Communications Technology Planning \& Evaluation).

\bibliography{references, custom}

\appendix
%\section{Appendix}
\section{Full Example}
\begin{figure*}[p]
  \centering
  \begin{minipage}{\linewidth}
  \begin{userprompt}{Input Prompt}
    Subreddit Info:
    \vspace{0.5\baselineskip}
    \begin{headerbox}{Subreddit: r/santamonica: Welcome to Santa Monica, California}
    Anything related to Santa Monica, where the rent is high but the temperatures aren't.
    \end{headerbox}
    \vspace{0.5\baselineskip}
    Rules:
    \vspace{0.5\baselineskip}
    \begin{headerbox}{Rule 1:  Santa Monica-related only}
    Please keep posts to Santa Monica-specific issues. Venice and West LA issues can be posted in /r/LosAngeles
    \end{headerbox}
    \vspace{0.5\baselineskip}
    \begin{headerbox}{Rule 2:  Respect other redditors}
    Respect other r/SantaMonica users, as both individuals and as groups. If you disagree with someone here, do so gracefully without personal attacks. And remember, members of Santa Monica's unhoused population use this subreddit, too.
    \end{headerbox}
    \vspace{0.5\baselineskip}
    \begin{headerbox}{Rule 3:  No spam}
    We allow local businesses to share here if they're offering specials to the r/SantaMonica community. If multiple posts are submitted within short succession on the same topic (e.g. ``earthquake!''), they will be consolidated down to 1.
    \end{headerbox}
    \vspace{0.5\baselineskip}
    \begin{headerbox}{Rule 4:  Appealing bans}
    The moderators of this Subreddit seek to make this a place for civil discourse on Santa Monica and new users are held to a high standards for civility and quality of contributions. Mods enforce the rules using bans that allow people to read the rules. If you were banned and would like to continue participating in this forum, let us know that you've read the rules and why you'd like to continue participating in this forum versus other online forums in your ban appeal.
    \end{headerbox}
    \vspace{0.5\baselineskip}
    \begin{headerbox}{Rule 5:  No low-effort posts}
    Sometimes low-quality posts get caught in the subreddit's filters. These are removed by auto-moderator and removal is confirmed by moderators. Include an adequate and accurate description in your post; provide sources if asked by the moderators. Don't use this Subreddit as your personal research assistant if you're not interested in searching the history.
    \end{headerbox}
    \vspace{0.5\baselineskip}
    \begin{headerbox}{Rule 6:  No calling people paid shills just because they disagree with you.}
    Accusing people of being paid shills with no proof just because they disagree with you can and will result in bans.
    \end{headerbox}
    \vspace{0.5\baselineskip}
    \begin{headerbox}{Rule 7:  No hateful speech directed at other people, including about the homeless.}
    If you want to say something disgusting about homeless people, NextDoor is always there for you.
    \end{headerbox}
    \par
    \vspace{0.5\baselineskip}
    Submission:
    \vspace{0.5\baselineskip}
    \begin{headerbox}{Homeowners association sues City and Fairmont Miramar over hotel redevelopment approval\\USER1, Tue, Dec 15, 2020, 1:37PM}
    \url{https://www.smdp.com/homeowners-association-sues-city-and-fairmont-miramar-over-hotel-redevelopment-approval/199916}
    \begin{center}
      \includegraphics[width=0.25\textwidth]{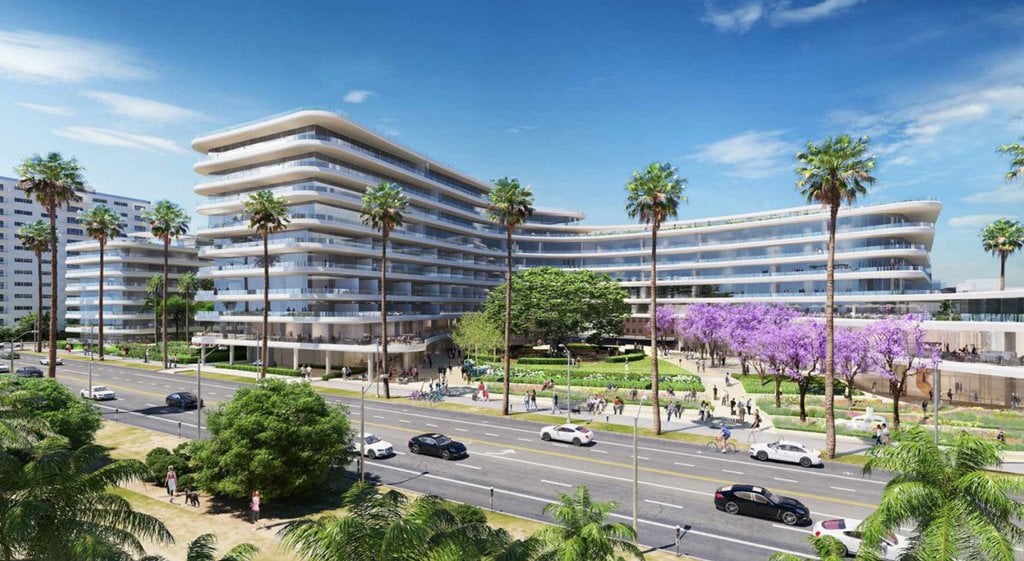}
    \end{center}
    \end{headerbox}
    \vspace{0.5\baselineskip}
    Discussion:
    \vspace{0.5\baselineskip}
    \begin{headerbox}{Comment 1: USER1, Tue, Dec 15, 2020, 1:38PM}
    And people wonder why everything is so expensive here. This redevelopment took 10 \textbf{years} to get approved and they're \textit{still} not out of the woods on getting held up by NIMBY lawsuits. These risks get priced in on any development project and it's basically a guarantee a lot of would-be projects never get off the ground in the first place because they know they're risking interminable timelines like this.
    \end{headerbox}
    \vspace{0.5\baselineskip}
    \begin{headerbox}{Comment 2: USER2, Tue, Dec 15, 2020, 1:48PM}
    Shrug.

    How many more empty storefronts does Santa Monica need?

    Third Street Promenade has thousands of square feet of empty available commercial space. The former Barnes \& Noble, the former Banana Republic. Dozens of empty storefronts a block away.

    This is a pointless multi-million dollar development that no one is asking for.

    Meanwhile thousands of unhoused people on the streets\ldots
    \end{headerbox}
    \vspace{0.5\baselineskip}
    \begin{headerbox}{Comment 3: USER3, Tue, Dec 15, 2020, 4:45PM}
    This is a hotel. Hotels bring more patrons for our local businesses. They also produce a ton of tax revenue.
    \end{headerbox}
    \vspace{0.5\baselineskip}
    \begin{headerbox}{Comment 4 [TARGET COMMENT]: USER4, Mon, Dec 21, 2020, 3:58PM}
    You must work for Dell. He's been ripping off Santa Monica for years by avoiding taxes on the Fairmont Miramar. Look it up. He's a right wing a-hole
    \end{headerbox}
    \vspace{\baselineskip}
    Question: Does the [TARGET COMMENT] violate a rule?
    \vspace{0.5\baselineskip}

    (a)~Santa Monica-related only; (b)~No rules broken; (c)~Respect other redditors; (d)~Appealing bans; (e)~No calling people paid shills just because they disagree with you; (f)~No low-effort posts; (g)~No hateful speech directed at other people, including about the homeless; (h)~No spam
  \end{userprompt}
  \vspace{0.25\baselineskip}
  \begin{assistantprompt}{Model Response}
    \textbf{Reasoning trace:} Evaluating user responses. I think I'd choose option (c): ``respect other redditors'' because calling someone a ``right wing a-hole'' definitely feels like a personal attack. The question is about whether that behavior violates a rule, and it clearly does. So, saying option (c) is the best choice makes sense to me. It's really important to foster a respectful community, and this kind of language just doesn't fit that.

    \vspace{0.5\baselineskip}
    \textbf{Final answer:} (c) Respect other redditors. \incorrectmark
  \end{assistantprompt}
  \end{minipage}
  \caption{Full \textsc{PluRule} example with complete input prompt and reasoning trace from GPT-5.2 (high reasoning).
  The correct answer is (e) No calling people paid shills, since ``You must work for Dell'' is an unsupported accusation.
  The model instead locks onto the incivility (``right wing a-hole'') and picks (c) Respect other redditors.}
  \label{fig:example-datapoint-full}
\end{figure*}

Figure~\ref{fig:example-datapoint-full} expands the abridged datapoint in Figure~\ref{fig:example-datapoint} with all seven community rules, the complete discussion thread, and GPT-5.2's verbatim reasoning trace.

\section{Supplementary Methods}
\label{sec:app:methods}

For clustering subreddits and rules, we perform grid searches over UMAP and HDBSCAN parameters: n\_neighbors, n\_components, min\_cluster\_size, and min\_samples. 
We maximize DBCV (Density-Based Cluster Validity), which measures 
cluster separation and coherence.

For subreddits, optimal parameters yield DBCV = 0.448 with 24 clusters (66 items per cluster on average).
We treat the remaining subreddits as the `other' cluster (20.4\% noise).
For rules, optimal parameters yield DBCV = 0.569 with 26 clusters (94 items per cluster on average).
We treat the remaining rules as the `other' cluster (15.6\% noise).

%\paragraph{Cluster Labeling.}
We assign semantic labels to each cluster using Qwen3-30B-A3B-Thinking.
For each cluster, we prompt the model to generate 10 candidate labels and select the most common one via majority voting.
We then manually verify and refine the labels for consistency.
This produces 26 rule clusters (e.g., civility, self-promotion, spoilers, flair) and 24 subreddit clusters (e.g., politics, memes, gaming, trading).

\section{Extended Results}
\label{sec:app:results}

\paragraph{Recall Breakdowns.} Table~\ref{tab:results-mod-unmod} reports recall for violating and compliant comments across all models and context levels.

\paragraph{Weighted Accuracy.} If we were to weigh violating comments more heavily than compliant ones, the results are worse across all models.
Table~\ref{tab:results-across-contexts-weighted-2-1} reports a 2:1 weighted accuracy that prioritizes violation detection over compliance: every model's score is lower than in the symmetric Table~\ref{tab:results-across-contexts} --- e.g., GPT-5.2 (high reasoning) with full context drops from 57.6\% to 52.6\% --- because models consistently recall compliant comments better than violating ones.

\begin{figure*}
\centering
\includegraphics[width=\textwidth]{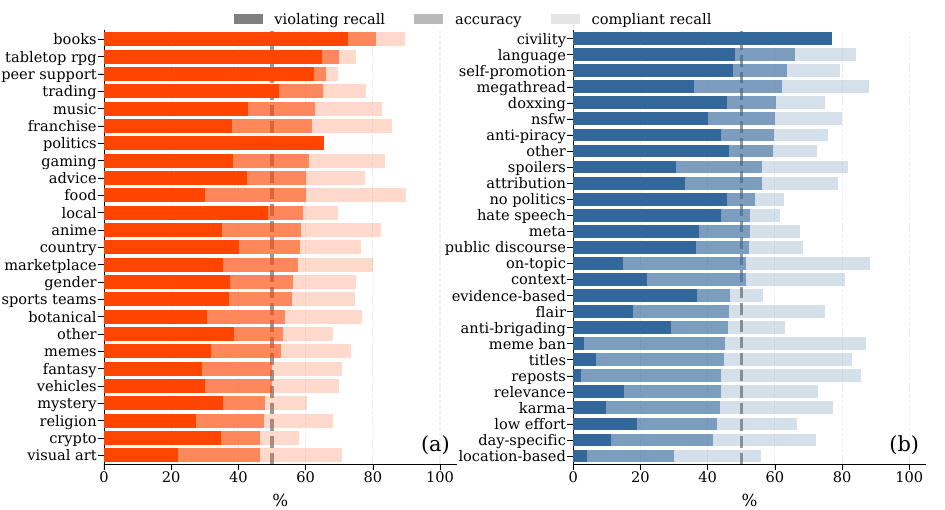}
\caption{GPT-5.2 (high reasoning) recall and accuracy with full context by (a)~subreddit cluster and (b)~rule cluster on the test set. Stacked bars show violating and compliant recall. Bars sorted by accuracy. Dashed lines indicate the 50\% baseline for accuracy.}
\label{fig:cluster-stacked}
\end{figure*}

\paragraph{GPT-5.2 Analysis.} Figure~\ref{fig:cluster-stacked} shows recall by subreddit and rule clusters for GPT-5.2 (high reasoning) with full context.

\begin{figure*}
  \centering
  \includegraphics[width=\textwidth]{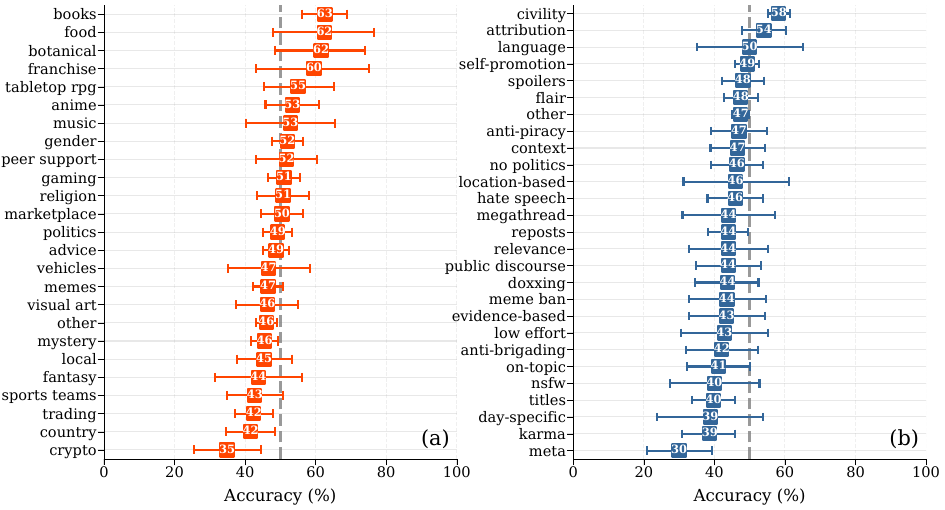}
  \caption{Accuracy for Qwen3-VL-4B (instruct) with full context by (a)~subreddit clusters and (b)~rule clusters. Error bars show 95\% CI. Dashed lines indicate the baseline.}
  \label{fig:qwen3-vl-4b-accuracy}
\end{figure*}

\begin{figure*}
  \centering
  \includegraphics[width=\textwidth]{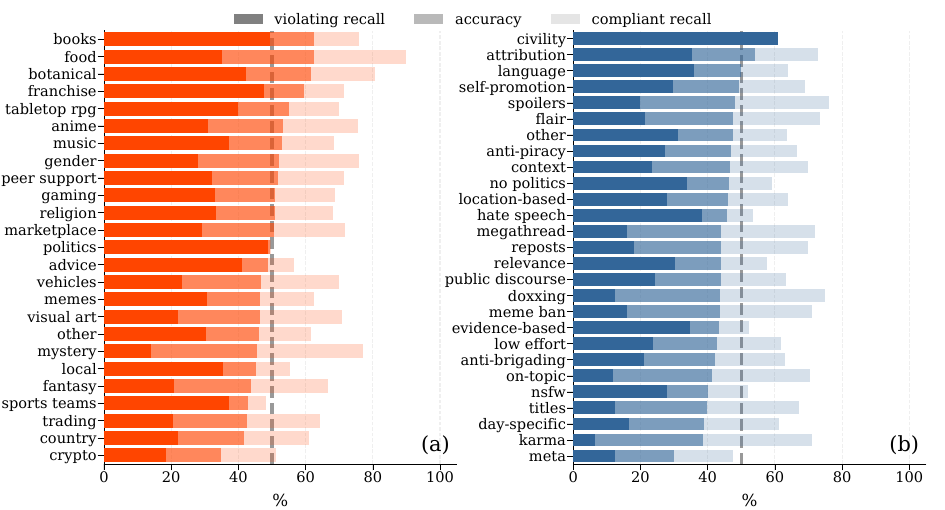}
  \caption{Qwen3-VL-4B (instruct) recall and accuracy with full context by (a)~subreddit cluster and (b)~rule cluster. Stacked bars show violating and compliant recall. Dashed lines indicate the 50\% baseline.}
  \label{fig:qwen3-vl-4b-stacked}
\end{figure*}

\begin{figure*}
  \centering
  \includegraphics[width=\textwidth]{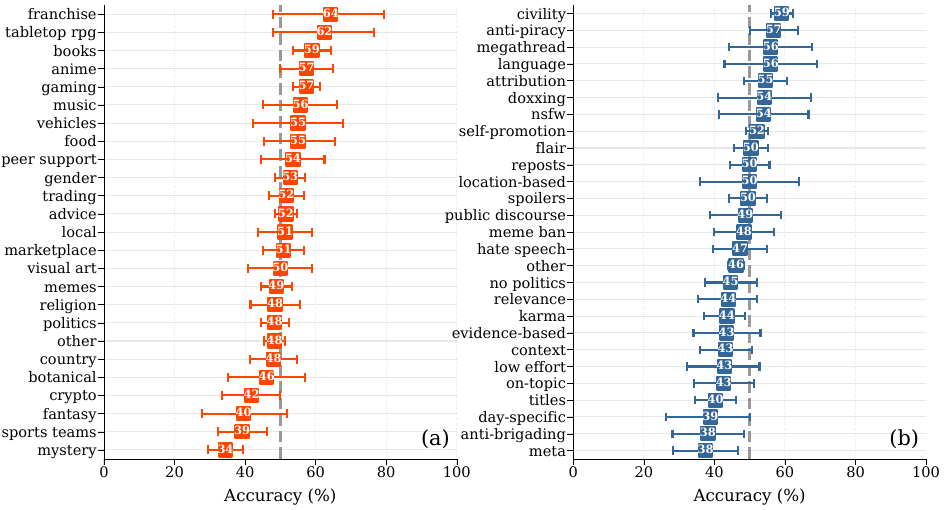}
  \caption{Accuracy for Qwen3-VL-8B (instruct) with full context by (a)~subreddit clusters and (b)~rule clusters. Error bars show 95\% CI. Dashed lines indicate the baseline.}
  \label{fig:qwen3-vl-8b-accuracy}
\end{figure*}

\begin{figure*}
  \centering
  \includegraphics[width=\textwidth]{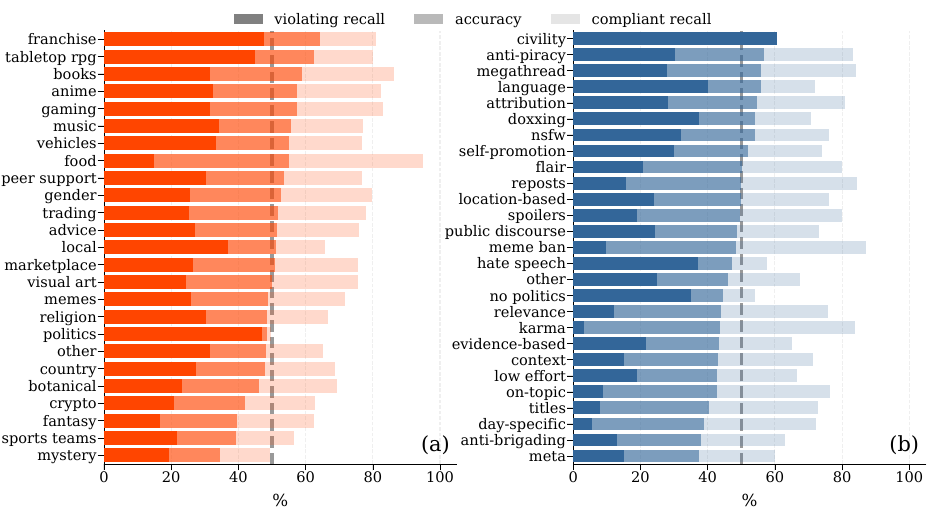}
  \caption{Qwen3-VL-8B (instruct) recall and accuracy with full context by (a)~subreddit cluster and (b)~rule cluster. Stacked bars show violating and compliant recall. Dashed lines indicate the 50\% baseline.}
  \label{fig:qwen3-vl-8b-stacked}
\end{figure*}

\begin{figure*}
  \centering
  \includegraphics[width=\textwidth]{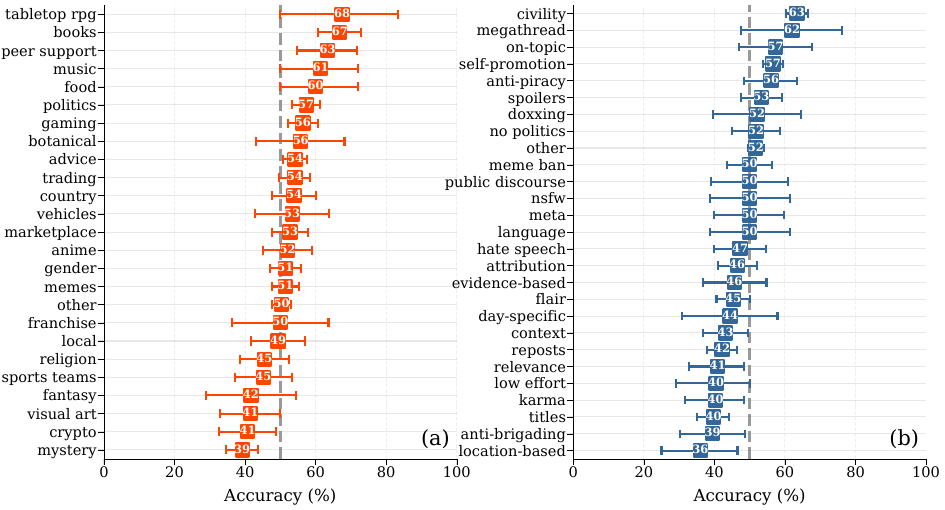}
  \caption{Accuracy for Qwen3-VL-30B (instruct) with full context by (a)~subreddit clusters and (b)~rule clusters. Error bars show 95\% CI. Dashed lines indicate the baseline.}
  \label{fig:qwen3-vl-30b-accuracy}
\end{figure*}

\begin{figure*}
  \centering
  \includegraphics[width=\textwidth]{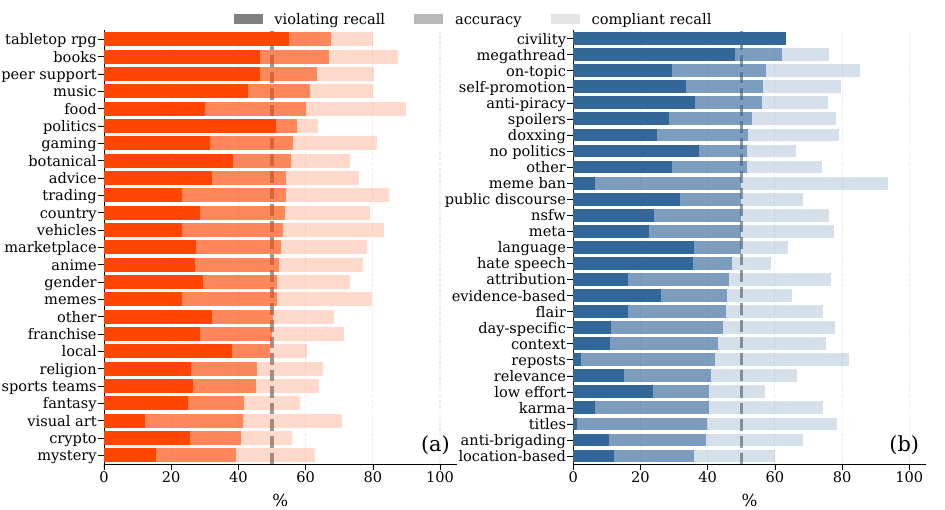}
  \caption{Qwen3-VL-30B (instruct) recall and accuracy with full context by (a)~subreddit cluster and (b)~rule cluster. Stacked bars show violating and compliant recall. Dashed lines indicate the 50\% baseline.}
  \label{fig:qwen3-vl-30b-stacked}
\end{figure*}

\paragraph{Qwen3-VL Analysis.} For each Qwen3-VL model (4B, 8B, 30B), we provide two figures: accuracy by subreddit and rule clusters (Figures~\ref{fig:qwen3-vl-4b-accuracy},~\ref{fig:qwen3-vl-8b-accuracy},~\ref{fig:qwen3-vl-30b-accuracy}) and recall breakdowns (Figures~\ref{fig:qwen3-vl-4b-stacked},~\ref{fig:qwen3-vl-8b-stacked},~\ref{fig:qwen3-vl-30b-stacked}).

\begin{figure*}
  \centering
  \includegraphics[width=\textwidth]{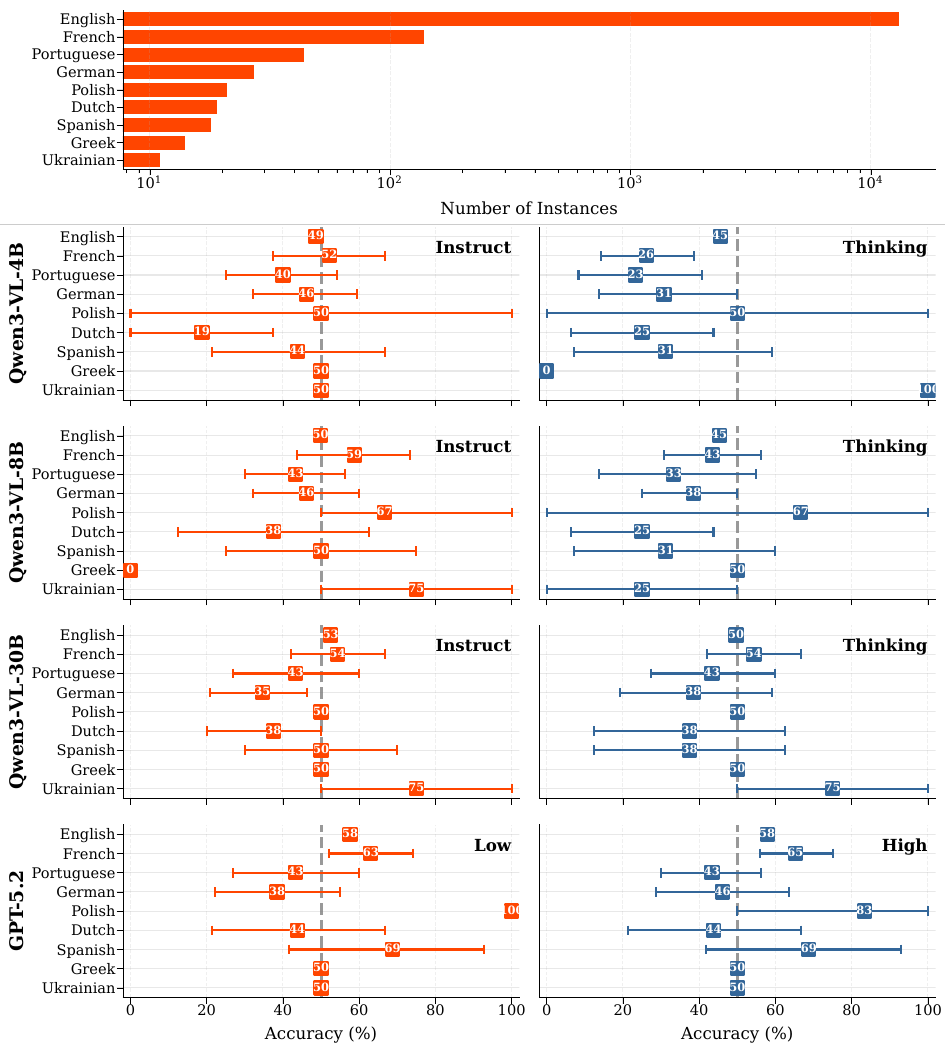}
  \caption{Language analysis across all evaluated models with full context. Top: distribution of PluRule instances across 9 languages. Bottom: per-language accuracy for Qwen3-VL-4B/8B/30B (instruct vs.\ thinking) and GPT-5.2 (low vs.\ high reasoning). Error bars show 95\% CI. Dashed lines indicate the 50\% baseline.}
  \label{fig:language-grid}
\end{figure*}

\paragraph{Language Analysis.} Figure~\ref{fig:language-grid} shows the distribution of PluRule instances across 9 languages and per-language accuracy for all evaluated models (Qwen3-VL-4B/8B/30B instruct and thinking; GPT-5.2 low and high reasoning) with full context.

\begin{figure*}
  \centering
  \includegraphics[width=\textwidth]{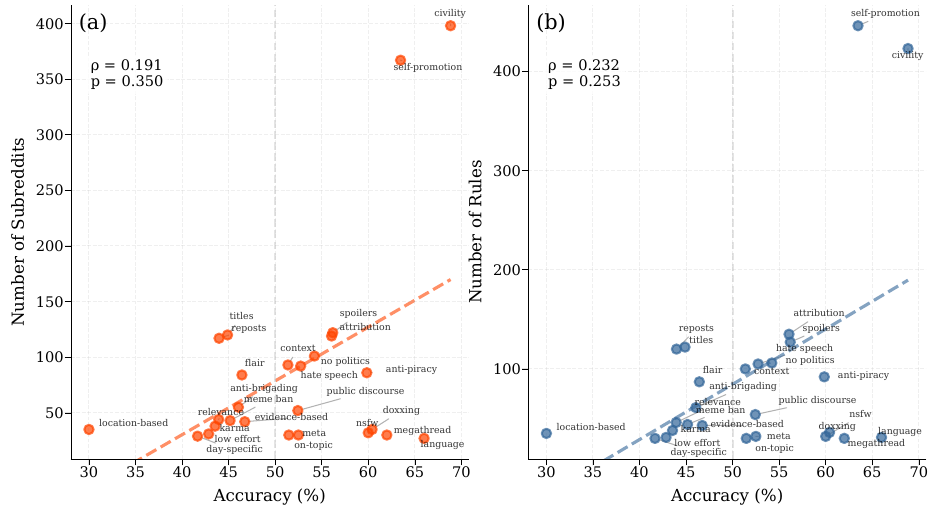}
  \caption{GPT-5.2 (high reasoning) accuracy with full context compared against (a)~the number of subreddits containing rules in each cluster and (b)~the number of rules per cluster. Dashed lines indicate the 50\% baseline.}
  \label{fig:rule-cluster-correlation}
\end{figure*}

\begin{figure*}
  \centering
  \includegraphics[width=\textwidth]{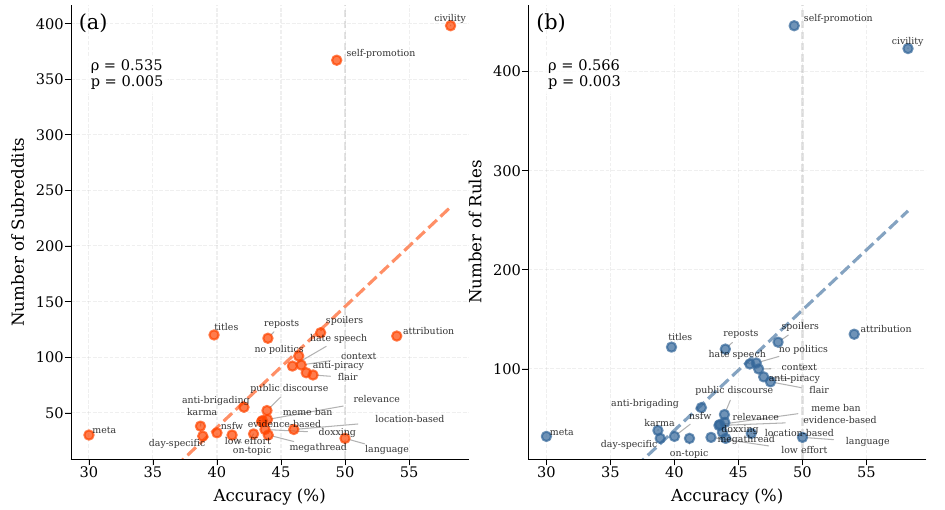}
  \caption{Qwen3-VL-4B (instruct) accuracy with full context compared against (a)~the number of subreddits containing rules in each cluster and (b)~the number of rules per cluster. Dashed lines indicate the 50\% baseline.}
  \label{fig:rule-cluster-correlation-4b}
\end{figure*}

\begin{figure*}
  \centering
  \includegraphics[width=\textwidth]{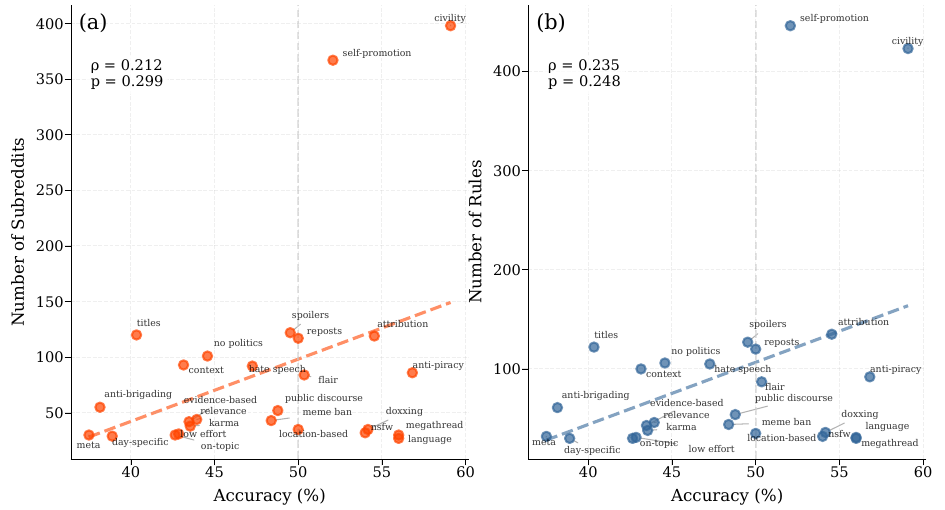}
  \caption{Qwen3-VL-8B (instruct) accuracy with full context compared against (a)~the number of subreddits containing rules in each cluster and (b)~the number of rules per cluster. Dashed lines indicate the 50\% baseline.}
  \label{fig:rule-cluster-correlation-8b}
\end{figure*}

\begin{figure*}
  \centering
  \includegraphics[width=\textwidth]{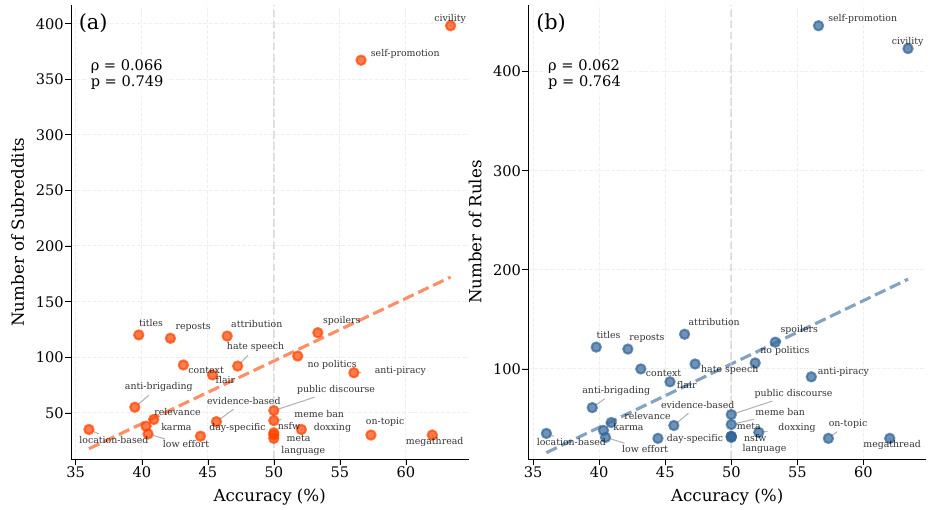}
  \caption{Qwen3-VL-30B (instruct) accuracy with full context compared against (a)~the number of subreddits containing rules in each cluster and (b)~the number of rules per cluster. Dashed lines indicate the 50\% baseline.}
  \label{fig:rule-cluster-correlation-30b}
\end{figure*}

\paragraph{Universality Correlation.} To examine whether models perform better on universal rule clusters, we measure universality in two ways: (a)~the number of subreddits containing at least one rule in the cluster, and (b)~the number of member rules in the cluster.
Figures~\ref{fig:rule-cluster-correlation},~\ref{fig:rule-cluster-correlation-4b},~\ref{fig:rule-cluster-correlation-8b}, and~\ref{fig:rule-cluster-correlation-30b} plot accuracy against both measures.
Spearman correlations are weak ($\rho=0.11$--$0.40$), with only Qwen3-VL-8B reaching significance.
However, civility and self-promotion --- the most universal by both measures --- consistently rank among the highest-accuracy clusters, with the exception of self-promotion for Qwen3-VL-4B.

\begin{table*}
  \centering
  \setlength{\tabcolsep}{4pt}
  \begin{tabular}{llllllll}
  \toprule
  \textbf{Model} & \textbf{Variant} & \textbf{Metric} & \textbf{Comment} & \textbf{+Discussion} & \textbf{+Submission} & \textbf{+User} & \textbf{+Images} \\
  \midrule
  \multirow{6}{*}{Qwen3-VL-4B} & \multirow{3}{*}{Instruct} & V. Recall & 24.1 & 26.4 {\tiny (+2.3)} & 29.8 {\tiny (+3.4)} & 30.7 {\tiny (+0.9)} & \textbf{32.6} {\tiny (+1.9)} \\
   & & C. Recall & \textbf{75.1} & 72.0 {\tiny (-3.1)} & 66.7 {\tiny (-5.3)} & 67.1 {\tiny (+0.4)} & 64.3 {\tiny (-2.8)} \\
   & & Accuracy & \textbf{49.6} & 49.2 {\tiny (-0.4)} & 48.3 {\tiny (-0.9)} & 48.9 {\tiny (+0.6)} & 48.4 {\tiny (-0.5)} \\
  \cmidrule(lr){2-8}
   & \multirow{3}{*}{Thinking} & V. Recall & 24.3 & 26.1 {\tiny (+1.8)} & 27.9 {\tiny (+1.8)} & \textbf{30.3} {\tiny (+2.4)} & 29.0 {\tiny (-1.3)} \\
   & & C. Recall & 50.5 & 53.5 {\tiny (+3.0)} & \textbf{62.0} {\tiny (+8.5)} & 59.8 {\tiny (-2.2)} & 60.9 {\tiny (+1.1)} \\
   & & Accuracy & 37.4 & 39.8 {\tiny (+2.4)} & 44.9 {\tiny (+5.1)} & \textbf{45.0} {\tiny (+0.1)} & 45.0 {\tiny (+0.0)} \\
  \midrule
  \multirow{6}{*}{Qwen3-VL-8B} & \multirow{3}{*}{Instruct} & V. Recall & 27.9 & 27.8 {\tiny (-0.1)} & 27.7 {\tiny (-0.1)} & 29.6 {\tiny (+1.9)} & \textbf{29.9} {\tiny (+0.3)} \\
   & & C. Recall & \textbf{74.1} & 73.7 {\tiny (-0.4)} & 70.8 {\tiny (-2.9)} & 70.4 {\tiny (-0.4)} & 69.6 {\tiny (-0.8)} \\
   & & Accuracy & \textbf{51.0} & 50.7 {\tiny (-0.3)} & 49.2 {\tiny (-1.5)} & 50.0 {\tiny (+0.8)} & 49.8 {\tiny (-0.2)} \\
  \cmidrule(lr){2-8}
   & \multirow{3}{*}{Thinking} & V. Recall & 31.2 & 33.2 {\tiny (+2.0)} & 32.6 {\tiny (-0.6)} & \textbf{34.7} {\tiny (+2.1)} & 32.3 {\tiny (-2.4)} \\
   & & C. Recall & 49.3 & 54.6 {\tiny (+5.3)} & \textbf{61.8} {\tiny (+7.2)} & 58.6 {\tiny (-3.2)} & 57.5 {\tiny (-1.1)} \\
   & & Accuracy & 40.3 & 43.9 {\tiny (+3.6)} & \textbf{47.2} {\tiny (+3.3)} & 46.7 {\tiny (-0.5)} & 44.9 {\tiny (-1.8)} \\
  \midrule
  \multirow{6}{*}{Qwen3-VL-30B} & \multirow{3}{*}{Instruct} & V. Recall & 30.2 & 31.4 {\tiny (+1.2)} & 31.1 {\tiny (-0.3)} & \textbf{32.4} {\tiny (+1.3)} & 31.6 {\tiny (-0.8)} \\
   & & C. Recall & 70.1 & 70.5 {\tiny (+0.4)} & 71.1 {\tiny (+0.6)} & 72.4 {\tiny (+1.3)} & \textbf{72.9} {\tiny (+0.5)} \\
   & & Accuracy & 50.2 & 51.0 {\tiny (+0.8)} & 51.1 {\tiny (+0.1)} & \textbf{52.4} {\tiny (+1.3)} & 52.3 {\tiny (-0.1)} \\
  \cmidrule(lr){2-8}
   & \multirow{3}{*}{Thinking} & V. Recall & 38.1 & \textbf{41.0} {\tiny (+2.9)} & 40.0 {\tiny (-1.0)} & 41.0 {\tiny (+1.0)} & 40.3 {\tiny (-0.7)} \\
   & & C. Recall & 54.1 & 55.4 {\tiny (+1.3)} & 58.3 {\tiny (+2.9)} & 57.8 {\tiny (-0.5)} & \textbf{58.6} {\tiny (+0.8)} \\
   & & Accuracy & 46.1 & 48.2 {\tiny (+2.1)} & 49.1 {\tiny (+0.9)} & 49.4 {\tiny (+0.3)} & \textbf{49.5} {\tiny (+0.1)} \\
  \midrule
  \multirow{6}{*}{GPT-5.2} & \multirow{3}{*}{Low} & V. Recall & 40.1 & 41.2 {\tiny (+1.1)} & 40.8 {\tiny (-0.4)} & \textbf{42.1} {\tiny (+1.3)} & 41.7 {\tiny (-0.4)} \\
   & & C. Recall & 68.1 & 69.3 {\tiny (+1.2)} & 72.9 {\tiny (+3.6)} & 72.7 {\tiny (-0.2)} & \textbf{73.1} {\tiny (+0.4)} \\
   & & Accuracy & 54.1 & 55.3 {\tiny (+1.2)} & 56.8 {\tiny (+1.5)} & \textbf{57.4} {\tiny (+0.6)} & \textbf{57.4} {\tiny (+0.0)} \\
  \cmidrule(lr){2-8}
   & \multirow{3}{*}{High} & V. Recall & 40.9 & 42.8 {\tiny (+1.9)} & 42.2 {\tiny (-0.6)} & \textbf{43.3} {\tiny (+1.1)} & 42.6 {\tiny (-0.7)} \\
   & & C. Recall & 69.2 & 69.7 {\tiny (+0.5)} & 72.4 {\tiny (+2.7)} & 72.1 {\tiny (-0.3)} & \textbf{72.7} {\tiny (+0.6)} \\
   & & Accuracy & 55.0 & 56.2 {\tiny (+1.2)} & 57.3 {\tiny (+1.1)} & \textbf{57.7} {\tiny (+0.4)} & 57.6 {\tiny (-0.1)} \\
  \midrule
  \multirow{3}{*}{Baseline} & & V. Recall & \multicolumn{5}{c}{0.0} \\
   & & C. Recall & \multicolumn{5}{c}{100.0} \\
   & & Accuracy & \multicolumn{5}{c}{50.0} \\
  \bottomrule
  \end{tabular}
  \caption{Violating recall, compliant recall, and accuracy (\%) across different models and contexts on the test set. Numbers in parentheses indicate differences compared to values in the previous column. Best-performing contexts for each model variant are highlighted in bold. All values have 95\% CI of at most $\pm 1.9\%$.}
  \label{tab:results-mod-unmod}
  \end{table*}

\begin{table*}[t]
  \centering
  \setlength{\tabcolsep}{3.75pt}
  \begin{tabular}{lllllllll}
  \toprule
  \textbf{Models} & \multicolumn{2}{c}{\textbf{Qwen3-VL-4B}} & \multicolumn{2}{c}{\textbf{Qwen3-VL-8B}} & \multicolumn{2}{c}{\textbf{Qwen3-VL-30B}} & \multicolumn{2}{c}{\textbf{GPT-5.2}} \\
  \cmidrule(lr){2-3} \cmidrule(lr){4-5} \cmidrule(lr){6-7} \cmidrule(lr){8-9}
  \textbf{Variants} & \multicolumn{1}{c}{Instruct} & \multicolumn{1}{c}{Thinking} & \multicolumn{1}{c}{Instruct} & \multicolumn{1}{c}{Thinking} & \multicolumn{1}{c}{Instruct} & \multicolumn{1}{c}{Thinking} & \multicolumn{1}{c}{Low} & \multicolumn{1}{c}{High} \\
  \midrule
  Comment Only & 41.1 & 33.1 & \textbf{43.3} & 37.3 & 43.5 & 43.4 & 49.4 & 50.3 \\
  \quad +Discussion & 41.6 {\tiny (+0.5)} & 35.2 {\tiny (+2.1)} & 43.1 {\tiny (-0.2)} & 40.3 {\tiny (+3.0)} & 44.5 {\tiny (+1.0)} & 45.8 {\tiny (+2.4)} & 50.6 {\tiny (+1.2)} & 51.8 {\tiny (+1.5)} \\
  \quad\quad +Submission & 42.1 {\tiny (+0.5)} & 39.2 {\tiny (+4.0)} & 42.0 {\tiny (-1.1)} & 42.3 {\tiny (+2.0)} & 44.5 {\tiny (+0.0)} & 46.1 {\tiny (+0.3)} & 51.5 {\tiny (+0.9)} & 52.3 {\tiny (+0.5)} \\
  \quad\quad\quad +User & 42.8 {\tiny (+0.7)} & \textbf{40.1} {\tiny (+0.9)} & 43.2 {\tiny (+1.2)} & \textbf{42.7} {\tiny (+0.4)} & \textbf{45.7} {\tiny (+1.2)} & \textbf{46.6} {\tiny (+0.5)} & \textbf{52.3} {\tiny (+0.8)} & \textbf{52.9} {\tiny (+0.6)} \\
  \quad\quad\quad\quad +Images & \textbf{43.1} {\tiny (+0.3)} & 39.6 {\tiny (-0.5)} & 43.2 {\tiny (+0.0)} & 40.7 {\tiny (-2.0)} & 45.4 {\tiny (-0.3)} & 46.4 {\tiny (-0.2)} & 52.2 {\tiny (-0.1)} & 52.6 {\tiny (-0.3)} \\
  \midrule
  Baseline & \multicolumn{8}{c}{33.3} \\
  \bottomrule
  \end{tabular}
  \caption{Weighted accuracy (\%) with violating:compliant ratio of 2:1 across models and context levels on the test set. Numbers in parentheses show differences from the previous row. Best-performing contexts for each model variant are highlighted in bold. The no moderation baseline drops to 33.3\% under this weighting.}
  \label{tab:results-across-contexts-weighted-2-1}
\end{table*}

\end{document}